\documentclass[11pt,a4paper]{article}
\usepackage[hyperref]{naaclhlt2019}
\usepackage{times}
\usepackage{latexsym}
\usepackage{graphicx}
\usepackage{amsmath}

\usepackage{url}

\aclfinalcopy 
\def\aclpaperid{2023} 

\renewcommand{\vec}[1]{\mathbf{#1}}

\title{Simplified Neural Unsupervised Domain Adaptation}

\author{Timothy Miller \\
  Computational Health Informatics Program \\
  Boston Children's Hospital \\
  Harvard Medical School \\
  {\tt timothy.miller@childrens.harvard.edu} }

\date{}

\begin{document}
\maketitle
\begin{abstract}
Unsupervised domain adaptation (UDA) is the task of modifying a statistical model trained on labeled data from a source domain to achieve better performance on data from a target domain, with access to only unlabeled data in the target domain.
Existing state-of-the-art UDA approaches use neural networks to learn representations that can predict the values of subset of important features called ``pivot features.''
%
In this work, we show that it is possible to improve on these methods by jointly training the representation learner with the task learner, and examine the importance of existing pivot selection methods.
%
\end{abstract}

\section{Introduction}
Unsupervised domain adaptation (UDA) is the task of modifying a statistical model trained on labeled data from a source domain to achieve better performance on data from a target domain, without access to any labeled data in the target domain.
Supervised domain adaptation methods can obtain excellent performance from a small number of labeled examples in the target domain~\cite{daume2007frustratingly}, but UDA is attractive in cases where annotation requires specialized expertise or the number of meaningfully different sub-domains is large (e.g., both are true for clinical NLP).
%

Structural correspondence learning~\cite{Blitzer2006DomainLearning} (SCL) is one widely-used method for UDA in natural language processing.
The key idea in SCL is that a subset of features, believed to be predictive across domains, are selected as \emph{pivot features}.
For each selected pivot feature, SCL creates an auxiliary classification task of predicting the value of that feature in an instance, given the values of all the non-pivot features for that instance.
The auxiliary classifiers therefore learn important cross-domain information about the structure of the feature space, which the SCL algorithm uses to create an augmented representation that aligns features from different domains (further details in Section~\ref{sect:bg}).

Meanwhile, recent advances in neural network learning have shown that training regimens that jointly consider evidence from multiple sources can improve performance -- both multi-task learning~\cite{Sgaard2016} and fine tuning~\cite{howard2018universal,devlin_bert:_2018}.
However, existing SCL-based methods treat the representation learning and task learning as separate tasks, so the parameters of the representation learning machinery are fixed before training for the downstream task.
Jointly learning the representation- and task-specific parameters can potentially allow a learning algorithm to find representations that are better suited for the task.

%

%
In this work, we describe a new UDA algorithm that is trained to jointly maximize two objectives: the primary supervised task in the source domain, and a pivot feature reconstruction task that can be trained on unlabeled data.
We also explore the importance of pivot feature selection to this algorithm, in experiments that quantitatively and qualitatively examine the quality of existing pivot selection methods. 
We find that our joint neural approach to SCL improves unsupervised domain adaptation substantially on a standard sentiment classification task.
Our results also show that while existing pivot selection methods perform well, they are below an oracle-provided ceiling for many source-target pairs for the sentiment classification task we examine.
%

\section{Background}
\label{sect:bg}
This work builds off of existing work in unsupervised domain adaptation, starting with Blitzer's work on structural correspondence learning (SCL)~\cite{Blitzer2006DomainLearning,Blitzer2007BiographiesClassification}.
In the UDA task setup, one is given two datasets, the source $D_S = \{ X_s, y_s \}$, with labels for each instance, and the target $D_T = \{ X_t \}$, with unlabeled instances only.
The goal of UDA is to learn a function $f_u(X_s, y_s, X_t) $ that improves on the classification performance over a function $f_l(X_s, y_s)$ when applied to new data drawn from the target distribution.

SCL is essentially a representation learning algorithm that works by creating a number of auxiliary classification tasks from the unlabeled source and target training instances (inspired by \citealt{ando2005framework}).
First, a set of $p$ \emph{pivot features} are selected, intended (in Blitzer's words) to be ``features which behave in the same way for discriminative learning in both domains.''
Then, SCL creates $p$ auxiliary tasks of predicting the value of pivot features in an instance given the non-pivot features in the instance.
The weights of these linear classifiers are then aligned as columns in a matrix $W$, and the $k$ left singular vectors are chosen from the singular value decomposition $W = U \Sigma V^{\top}$ to reduce its dimensionality, leading to a projection matrix $\theta \in R^{n \times k}$ that maps instances from the original feature space into the learned space.
Most practical implementations find the best performance of SCL is obtained when projected features are concatenated with the original feature space; for some tasks and datasets other combinations have been tested and proved superior~\cite{Sapkota2016DomainLearning}.

Recently, neural-network-based domain adaptation algorithms have been successful, including domain adversarial methods~\cite{ganin2016domain} and auto-encoder-based methods~\cite{glorot2011domain,chen2014marginalized}.
However, a neural version of SCL still obtains near state-of-the-art performance~\cite{ziser_reichart17}.
In that work, the AE-SCL system uses a multi-layer perceptron to replace the SVD for learning the feature projection.
This network takes non-pivot features as input, has one hidden layer, and predicts the value of the pivot features at the output layer.
Since it obtains supervision from the values of features, it can be trained on unlabeled instances from the source and target domains.
To train for the downstream sentiment classification task, the source instances are first passed into the trained representation learning network, and the values of the hidden layer are considered an additional set of features.
These features are combined with all the original features, and the authors use a logistic regression classifier for the final sentiment classifier.

One standard corpus used to develop new domain adaptation algorithms is the Amazon sentiment analysis dataset.\footnote{\url{https://www.cs.jhu.edu/~mdredze/datasets/sentiment/}}
This corpus was created by~\citet{Blitzer2007BiographiesClassification}, but we use the version included in the software release from~\citet{ziser_reichart17}\footnote{\url{https://github.com/yftah89/Neural-SCL-Domain-Adaptation}}, along with their pre-processing steps, for ease of comparison with their results.
This dataset contains reviews from four product categories on Amazon.com -- books, DVDs, electronics, and kitchen appliances.
Reviews are mapped to binary categories: \emph{positive} if the review assigns the product $>$3 stars (out of 5) and \emph{negative} if it assigns the product $<$ 3 stars.
This dataset also contains additional unlabeled instances for each category, used for training the pivot predictor.

\section{Methods}
The current work has two motivating factors.
First, we would like to improve the performance of SCL using joint training.
%
Existing SCL-based methods are successful in treating pivot prediction as a pre-training phase, but joint training may improve UDA by allowing the network to find representations that are equally good at pivot reconstruction but better for downstream task performance.
%
Second, we would like to evaluate the quality of pivot selection methods and explore whether this step might be eliminated to simplify SCL.
%
%
%

We focus on feature-based UDA methods, as opposed to approaches that rely on embeddings~(e.g., \citealp{barnes2018domain,ziser_reichart18}), since our primary interest is in improving existing models developed with a feature engineering approach.
Such methods allow us to quickly adapt a number of different models to new datasets (e.g., for already-existing NLP pipeline software), rather than engineering new neural models from scratch for each of the pipeline tasks.
For that reason, we compare to the AE-SCL model of \citeauthor{ziser_reichart17}, rather than their subsequent models that take embeddings as input.
In any case, we show that with some tuning the AE-SCL model can obtain state-of-the-art performance for many pairs.

\begin{figure}[t]
\centering
\includegraphics[scale=0.38]{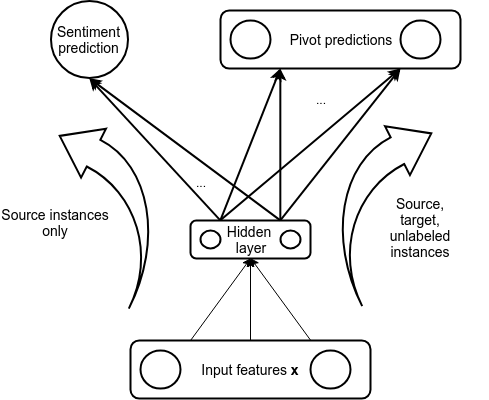}
\caption{Network architecture for unsupervised domain adaptation.}
\label{fig:model}
\end{figure}

\subsection{Joint Neural Structural Correspondence Learning}
\label{sect:joint}
%
%
Figure~\ref{fig:model} graphically depicts our proposed joint model.
The input to the model $\vec{x} \in \mathcal{R}^{n}$ is the set of all features extracted from the text -- to compare with~\citet{ziser_reichart17} we use unigrams and bigrams, extracted using scikit-learn~\cite{pedregosa2011scikit}.
We experimented with a few different hidden layer sizes, and settled on $d=2000$ -- this balances the need of the representation to predict more output variables than the AE-SCL method with run-time constraints.
%
The representation is generated with 
$h(\vec{x}) = ReLU(W_h \vec{x})$, for $W_h \in \mathcal{R}^{d \times n}$.
The task prediction is $f_{task}(\vec{x}) = Sigmoid(W_t h(\vec{x}))$ ($ W_t \in \mathcal{R}^{1 \times d}$) and the pivot prediction is $f_{pivot}(\vec{x}) = Sigmoid(W_p h(\vec{x}))$  ($ W_p \in \mathcal{R}^{p \times d}$).

%
The joint loss function for labeled source data $D_l$, all data $D_a$, and model parameters $\theta$ is:

\begin{equation}
    \begin{split}
    \mathcal{L}(D;\theta) = & \!\!\!\! \sum_{\vec{x},y \in D_l} BCE(f_{task}(\vec{x}),y) + \\
     & \!\!\!\!\!\!\!\! \lambda \sum_{\vec{x} \in D_a} BCE(f_{pivot}(\vec{x}), pivots(\vec{x})) + \\
     & \!\!\!\!\!\!\!\! \rho \mathcal{R}(\vec{\theta})
    \end{split}
\end{equation}

where $BCE$ is the binary cross-entropy loss, $\lambda$ controls the weight of pivot prediction loss, $pivots$ is a function that selects the indices from an instance that are the pivot features to be predicted, and $\rho$ is the weight of the regularization term $\mathcal{R}$.

%
%

To train this model, we alternate passing labeled source data and unlabeled data from the source and target domains into the network.
For the labeled data, the error term is the sum of task- and pivot-prediction tasks, while for the unlabeled data only the pivot-prediction loss is computed.

Training proceeds for 30 epochs, with mini-batch size of 50 instances, using the Adam optimizer~\cite{kingma2014adam} with learning rate $0.001$.
For the loss function weight, we used $\rho=0.1$ and $\lambda=100$.
We used held-out source data to compute validation loss after each epoch and selected the trained model with the lowest validation loss.

\subsection{Pivot Selection for Neural Systems}
One standard way of choosing pivot features is by calculating mutual information (MI) between the source features and labels, and selecting the features with the highest MI.
It is far from clear, however, that this technique is always optimal.
Earlier experiments with the POS tagging task~\cite{Blitzer2006DomainLearning} used feature frequency instead, and the extent of the correlation between frequency and MI for that task is not established.

Here, we attempt to provide some evidence about the quality of MI for the task of sentiment classification, using the classification pair of books to electronics. 
First, we wanted to rule out the null hypothesis that prediction of MI pivots is essentially a generic representation learning algorithm -- in other words, that a network learning structure between any sets of sufficiently common features may improve adaptation performance.
We modified Ziser et al.'s code to simply select random feature indices from the subset of those that occurred frequently enough to be pivot candidates.
With this setup, adaptation performance averaged $0.724$ across ten runs, well below their reported $0.744$, casting doubt on the null hypothesis.

Next, we want to examine the contention that features with high MI relative to source labels are general.
To do this, we simply compare the list of MI features used when the source is \emph{books} to those when the source is \emph{electronics}.
We find that, out of the 100 pivot features selected by MI in either cases, there is overlap of 26 features, some examples of which are shown in Table~\ref{table:mi_overlap} (left).
Table~\ref{table:mi_overlap} also shows a number of MI-selected pivots from the \emph{books} domain that are not general (middle), and then a set of features MI-correlated with the target domain that seem general (right).
These latter two columns are essentially precision and recall errors of the MI pivot selection algorithm.


Finally, we perform an oracle-based adaptation experiment, where we select the pivot feature indices using MI against the gold labels of the \emph{target} domain, but then proceed with training looking only at source labels, with results in Table~\ref{tab:results} discussed below.
%
%


\begin{table}
\begin{center}
\begin{tabular}[t]{c|c|c}
Shared features & Book (spec) & Elec (gen) \\
\hline
an excellent & writing & and great \\
best & care about & is perfect \\
bad & flat  & never buy \\
highly recommend & pages & not good \\
waste your & finish & poor \\
\hline
\end{tabular}
\caption{Example features selected as pivots using MI. \emph{Shared features} indicates those selected for both domains. \emph{Book (spec)} highlights pivots selected in the books domain that appear domain-specific. \emph{Elec (gen)} indicates high-MI pivots in the electronics domain that appear domain-general that are not selected when \emph{books} is the source domain.}
\label{table:mi_overlap}
\end{center}
\end{table}

\section{Evaluation}
\begin{table}[t]
    \centering
    \begin{tabular}{l|c|c|c|c}
    S$\rightarrow$T \!               & \!AE-SCL\! & \!AE-SCL$_{R}$\! & $Joint_{MI}$ & $Joint_{O}$ \\
    \hline
    B$\rightarrow$D     & 0.794 & \bf{0.812} & 0.808 & 0.84\\
    B$\rightarrow$E     & 0.744 & 0.761 & \bf{0.786} & 0.818\\
    B$\rightarrow$K     & 0.795 & \bf{0.814} & 0.809 & 0.831 \\
    D$\rightarrow$B     & 0.758 & 0.779 & \bf{0.807} & 0.808\\
    D$\rightarrow$E     & 0.763 & 0.763 & \bf{0.803} & 0.843 \\
    D$\rightarrow$K     & 0.8   & 0.821 & \bf{0.83} & 0.832 \\
    E$\rightarrow$B     & 0.701 & 0.701 & \bf{0.738} & 0.749 \\
    E$\rightarrow$D     & 0.732 & 0.748 & \bf{0.766} & 0.824 \\
    E$\rightarrow$K     & 0.848 & 0.848 & \bf{0.875} & 0.869 \\ 
    K$\rightarrow$B     & 0.742 & 0.731 & \bf{0.744} & 0.767 \\
    K$\rightarrow$D     & 0.743 & 0.754 & 0.752 & 0.828 \\
    K$\rightarrow$E     & 0.828 & 0.841 & \bf{0.856} & 0.851\\
    \hline
    Ave.                & 0.771 & 0.781 & \bf{0.798} & 0.821\\
    \hline
    \end{tabular}
    \caption{Summary of results. B=Books, D=Dvd, E=Electronics,K=Kitchen. AE-SCL=Reported results from \citet{ziser_reichart17}. AE-SCL$_{R}$=Replicated results from the same. $Joint_{MI}$=results from the joint model in Section~\ref{sect:joint}. $Joint_{O}$=results from the joint model using oracle MI pivot selection. Bold indicates a significant difference ($p < 0.05$) between AE-SCL$_{R}$ and $Joint_{MI}$ using Welch's one-tailed t-test.}
    \label{tab:results}
\end{table}

We follow the standard setup 
for the Amazon sentiment task, splitting each source dataset into 1600 training and 400 validation instances, and evaluating on the entire labeled target dataset for each pair.
We compare against two baselines: First, the reported results of \citeauthor{ziser_reichart17}, and second, our replication of their results using their code.
Our replication changed their code by replacing the stochastic gradient descent optimizer with Adam~\cite{kingma2014adam}, and increasing the training batch size from $1$ to $50$.
These changes were made to speed training runs during development; we found they produced better-than-reported results and include these superior results as an even stronger baseline.
We report results of two configurations of our joint learner.
The first configuration ($Joint_{MI}$) uses the MI between source labels and features to select 100 pivot features.
The second configuration ($Joint_{Oracle}$) is an oracle-informed system where we use the MI between \emph{target} labels and features to select pivot features, but only use source labels while training the network.
Both the AE-SCL$_R$ model and our $Joint_{MI}$ model were run for 10 iterations to minimize differences due to random initialization and to calculate significance statistics.

Table~\ref{tab:results} shows the results of our experiments.
First, we note that our replication of AE-SCL improves upon their reported results in 8 of 12 pairs, often by substantial margins, and is only worse in one pair (Kitchen$\rightarrow$Books).
Our $Joint_{MI}$ method is superior to the reported AE-SCL results in all pairs, 1.7 points (absolute) on average, and significantly better than the AE-SCL$_R$ in 9 of 12 pairs, using Welch's one-tailed t-test.
This is, to our knowledge, the best result on this task using a feature-based approach (i.e., excluding systems that use embeddings).
Despite constraining our system to adapting feature-based models, this result is competitive with the best-known result using a pure neural approach with embeddings as input, as \citet{ziser_reichart18} report an average accuracy of 0.804.
The $Joint_{Oracle}$ configuration shows that, despite the large gains of joint training, there is still significant improvement available with better pivot selection.

\section{Discussion and Conclusion}
Our results show that by jointly learning representations and task networks, UDA can be greatly improved over existing neural UDA methods.
We note that there are existing domain adaptation methods that use joint training with auxiliary tasks.
\citet{yu_learning_2016} use an auxiliary task of predicting whether a masked pivot word in a sentence is positive or negative sentiment, where they introduce a new technique to select pivots that still is based on correlations with source labels.
Our work is unique in showing that the standard task of mutual-information-selected pivot prediction is a high quality auxiliary task, though future work should explore whether their pivot selection algorithm is superior to MI in our joint model.
We also showed that existing neural UDA methods can be improved significantly with minor changes to the training regimen.
Finally, we show that mutual information pivot selection is quite far from the performance ceiling provided by oracle-based pivot selection.
%


This work evaluated on the widely-used Amazon sentiment dataset from~\citet{Blitzer2007BiographiesClassification}.
However, we believe that future work on domain adaptation should phase out the use of this dataset.\footnote{Thanks to the anonymous reviewer who made this argument which we found convincing. We credit them with these points while accepting the blame for any poor communication of these points.}
The test set for this setup is flawed in two important ways: first, it is artificially balanced with positive and negative reviews, when the problem is not actually balanced; it also has 3-star reviews removed, which is not a realistic test set setup without looking at labels.
For these reasons, we recommend that future work use different domain adaptation datasets.

The Pytorch implementation used to produce these results is publicly available.~\footnote{\url{https://github.com/tmills/Neural-SCL-Domain-Adaptation}}




\section*{Acknowledgments}
Research
reported in this publication was supported by the National Library Of Medicine of the National
Institutes of Health under Award Number R01LM012918. The content is solely the responsibility
of the authors and does not necessarily represent the official views of the National Institutes of
Health.

\bibliography{mendeley,zotero}
\bibliographystyle{acl_natbib}

\end{document}